\documentclass[a4paper,twoside]{article}

\usepackage{epsfig}
\usepackage{subcaption}
\usepackage{calc}
\usepackage{amssymb}
\usepackage{amstext}
\usepackage{amsmath}
\usepackage{amsthm}
\usepackage{multicol}
\usepackage{pslatex}
\usepackage{apalike}
\usepackage{algorithm2e}
\usepackage[bottom]{footmisc}
\usepackage{booktabs}
\usepackage{stfloats}
\usepackage{float}
\usepackage{url}
\usepackage{multirow}
\usepackage{graphicx}   
\usepackage{amsmath}    
\usepackage[normalem]{ulem} 
\usepackage{SCITEPRESS}     

\begin{document}

\title{TinyGraphEstimator: Adapting Lightweight Language Models for Graph Structure Inference}

\author{\authorname{Michal Podstawski
\orcidAuthor{0000-0003-1222-6894}
}
\affiliation{NASK National Research Institute, Warsaw, Poland}
\email{michal.podstawski@nask.pl}
}



\abstract{Graphs provide a universal framework for representing complex relational systems, and inferring their structural properties is a core challenge in graph analysis and reasoning. While large language models have recently demonstrated emerging abilities to perform symbolic and numerical reasoning, the potential of smaller, resource-efficient models in this context remains largely unexplored. This paper investigates whether compact transformer-based language models can infer graph-theoretic parameters directly from graph representations. To enable systematic evaluation, we introduce the TinyGraphEstimator dataset - a balanced collection of connected graphs generated from multiple random graph models and annotated with detailed structural metadata. We evaluate several small open models on their ability to predict key graph parameters such as density, clustering, and chromatic number. Furthermore, we apply lightweight fine-tuning using the Low-Rank Adaptation (LoRA) technique, achieving consistent improvements across all evaluated metrics. The results demonstrate that small language models possess non-trivial reasoning capacity over graph-structured data and can be effectively adapted for structural inference tasks through efficient parameter tuning.}

\onecolumn \maketitle \normalsize \setcounter{footnote}{0} \vfill

\section{\uppercase{Introduction}}

Graphs are a fundamental representation for modeling complex relationships in domains such as social networks, biology, and information systems. Accurately characterizing these networks requires estimating key graph-theoretic properties - such as connectivity, clustering, and chromatic number - that capture their structure and complexity. Traditional graph-learning approaches, including Graph Neural Networks and algorithmic estimators, achieve strong results but depend on task-specific architectures and explicit structural features.


To address this gap, we investigate whether small transformer-based language models can be adapted to estimate structural graph parameters from textual representations of graphs. In our setup, each graph is encoded in a deterministic edge-list format, which serves as a textual proxy for its topology. The models are trained to predict key graph-theoretic quantities from these structured inputs, enabling us to assess whether compact language models can approximate graph reasoning patterns when fine-tuned on explicit structural data, rather than relying on built-in algorithmic mechanisms or large model capacity.

To this end, we systematically evaluate several small, resource-efficient models on the task and fine-tune them using the Low-Rank Adaptation (LoRA) method. The results show that while these models exhibit limited capability in the zero-shot setting, fine-tuning leads to consistent and substantial improvements across all structural measures. These findings demonstrate that compact language models, when properly adapted, can acquire effective reasoning behavior over graph-structured data while maintaining high computational efficiency.

The main contributions of this paper are as follows:

\begin{itemize}
    \item We present a systematic evaluation of small, open language models on the task of inferring graph-theoretic parameters from structured graph representations.
    \item We demonstrate that LoRA-based fine-tuning yields consistent performance improvements, highlighting the potential of compact models for graph analysis tasks.
    \item We introduce the TinyGraphEstimator dataset, a balanced and publicly available collection of connected graphs annotated with key structural properties for reasoning evaluation.
\end{itemize}

\section{\uppercase{Related work}}

The task of predicting graph-theoretic properties has traditionally been addressed using specialized graph learning architectures such as Graph Neural Networks (GNNs)~\cite{kipf2017semi,hamilton2017inductive,wu2020comprehensive}. These models operate directly on adjacency structures to learn node embeddings and global representations that capture connectivity and topology. Numerous studies have explored the prediction of structural metrics including clustering coefficients, path lengths, and degree distributions from learned embeddings~\cite{you2018graph,dwivedi2020benchmarking}. However, while GNNs provide strong inductive biases for graph tasks, they often lack general reasoning ability outside the training distribution. Our work diverges from these approaches by treating the inference of graph properties as a language-based reasoning problem, where information is expressed textually and interpreted by small language models.

Recent advances in large language models (LLMs) have demonstrated impressive performance in tasks requiring symbolic reasoning, numerical computation, and pattern generalization~\cite{brown2020language,bubeck2023sparks}. Studies have shown that transformer-based architectures can internalize relational structures~\cite{wang2025reasoninglargelanguagemodels,schmitt2021modelinggraphstructurerelative,ying2021transformersreallyperformbad} and perform graph-related reasoning when provided with suitable prompts or structured input~\cite{huang2024llmseffectivelyleveragegraph,guo2023gpt4graphlargelanguagemodels,guo2025grapheditlargelanguagemodels}. Nevertheless, most prior work focuses on large-scale models with hundreds of billions of parameters, whose reasoning capabilities emerge from scale rather than explicit adaptation. In contrast, we investigate whether \emph{small} language models - under 4B parameters - can infer precise graph-theoretic quantities. This shifts the focus from emergent reasoning in massive models to lightweight, interpretable reasoning in compact architectures.

A growing body of work explores how language models can represent or reason about graphs through textual or hybrid encodings. These approaches encode adjacency information as token sequences, enabling LLMs to reconstruct or classify graph structures~\cite{tang2024graphgptgraphinstructiontuning,chen2025graphgenerativepretrainedtransformer,podstawski2025applyingtextembeddingmodels}. Other works have investigated relational reasoning benchmarks~\cite{hu2025rethinkingbenchmarkinglargelanguage,agrawal2024llmsperformstructuredgraph}, showing that LLMs can approximate certain structural statistics with appropriate context. Our approach builds on this foundation but emphasizes the systematic evaluation of \emph{structural parameter inference} - quantitative graph descriptors such as density, transitivity, and global efficiency - using small models specifically adapted for this purpose.

Fine-tuning language models for specialized reasoning tasks poses significant computational and memory challenges. Parameter-efficient adaptation techniques such as Low-Rank Adaptation (LoRA)~\cite{hu2021loralowrankadaptationlarge}, prefix tuning~\cite{li2021prefixtuningoptimizingcontinuousprompts}, and adapter-based methods~\cite{houlsby2019parameterefficienttransferlearningnlp} have emerged as scalable alternatives that modify only a small subset of parameters while preserving base model capabilities. These techniques have been successfully applied across a range of domains including code understanding~\cite{han2024parameterefficientfinetuninglargemodels}, reasoning~\cite{bi2024enhancingreasoningcapabilitiessmall}, and domain adaptation~\cite{lu2024finetuninglargelanguagemodels}. In our study, we employ LoRA to specialize small language models for structural graph inference, demonstrating consistent performance improvements across all measured graph parameters. This confirms that lightweight fine-tuning can effectively enhance reasoning abilities even in compact architectures.

In contrast to prior work that primarily investigates either graph representation learning or large-scale emergent reasoning, our research unifies both directions by systematically testing the graph reasoning capacity of small, adaptable language models. By combining structured graph representations with efficient fine-tuning, our approach provides new insights into how compact LMs internalize relational structure and approximate graph-theoretic computations.

\section{\uppercase{Proposed solution}}


The proposed approach evaluates and enhances the ability of small language models to infer a wide range of graph-theoretic parameters directly from structured graph representations. Each model operates on undirected, unweighted graphs whose topology is defined entirely by edge connectivity, and is prompted to predict a structured set of key structural properties that collectively describe graph topology and complexity. The inferred parameters include detailed degree-based statistics such as \textbf{minimum degree}, \textbf{mean degree}, \textbf{maximum degree}, and \textbf{degree standard deviation}, along with the global \textbf{density} metric describing overall connectivity. In addition to these fundamental attributes, we evaluate the models’ ability to estimate higher-order structural metrics including the total number of \textbf{triangles}, \textbf{average clustering coefficient}, \textbf{transitivity}, \textbf{average shortest path length}, \textbf{diameter}, \textbf{chromatic number}, and \textbf{global efficiency}. Taken together, these parameters capture both local and global aspects of graph organization, providing a comprehensive benchmark for assessing reasoning over graph structures.

In the first stage of our study, we test unmodified (plain) small language models to evaluate their intrinsic ability to perform numerical reasoning and pattern recognition over graph-structured data. This zero-shot evaluation establishes a baseline for each model’s inherent capacity to approximate structural graph measures from textual inputs. In the second stage, we apply lightweight fine-tuning using the Low-Rank Adaptation (LoRA) method, which enables efficient parameter adjustment without retraining the full model. This adaptation significantly improves performance across all evaluated parameters, demonstrating that even compact models can be effectively specialized for graph inference through minimal, targeted fine-tuning. The resulting framework provides a scalable and efficient solution for structural property estimation of graphs, highlighting the latent reasoning capabilities of small language models when guided by domain-specific adaptation.

\subsection{Models Selection}

To investigate the ability of small language models to infer structural properties of graphs, we selected a group of compact yet high-performing models that combine efficient deployment with strong reasoning capacity. Our goal was to identify models that are powerful enough to capture graph-theoretic patterns while remaining computationally accessible for systematic experimentation. Model selection was guided by comparative performance metrics reported on the \textit{LLM-Stats} leaderboard~\cite{llmstats_github_2024}, focusing on models with fewer than 4B parameters that demonstrate competitive reasoning accuracy across diverse benchmarks.

The final selection includes \textbf{Qwen-2.5-3B}~\cite{qwen2025qwen25technicalreport}, \textbf{Llama-3.2-3B}~\cite{grattafiori2024llama3herdmodels}, and \textbf{Phi-4-mini (4B)}~\cite{abdin2024phi4technicalreport}, all used in their \textit{Instruct} variants. These models were chosen because they consistently rank among the strongest sub-4B architectures in reasoning and comprehension tasks while maintaining full open availability for reproducible research. Their extended context capacity allows complete representations of graphs to be processed in a single inference window. Moreover, their compact architecture enables efficient fine-tuning and evaluation on standard hardware, making them ideal candidates for exploring how small-scale language models internalize and generalize graph-structural information.

For reference, we also evaluate two state-of-the-art LLMs - \textbf{GPT-4.1}~\cite{openai_gpt41_2024,openai2024gpt4technicalreport} and \textbf{DeepSeek-V3.2-Exp} (Non-Thinking Mode)~\cite{deepseek_docs_2025,deepseekai2025deepseekv3technicalreport} - to benchmark our compact models against the current frontier of LLM capability. While these systems exhibit strong general reasoning and broad-domain adaptability, our fine-tuned small models achieve superior accuracy on graph-structural inference, highlighting the efficiency and specialization benefits of lightweight adaptation.

\subsection{Dataset Construction}

To the best of our knowledge, no existing dataset directly supports the task of mapping raw graph structures to such a wide range of quantitative parameters. Therefore, we constructed a balanced synthetic dataset of connected graphs generated from three canonical random graph models: \textbf{Erdős–Rényi (ER)}, \textbf{Barabási–Albert (BA)}, and \textbf{Watts–Strogatz (WS)}. The ER model produces graphs where each possible edge occurs independently with a fixed probability, capturing purely random connectivity~\cite{erdos59a}. The BA model generates scale-free networks through preferential attachment, emphasizing hub formation and heavy-tailed degree distributions~\cite{barbasi_1999}. The WS model interpolates between regular lattices and random graphs, combining high clustering with short path lengths~\cite{watts_collective_1998}.

For each generated instance, the number of nodes \( n \) is uniformly sampled from the range \([20, 30]\), and connectivity is enforced by unbiased resampling until a connected instance is obtained, avoiding bias introduced by artificial augmentation. Each model contributes an equal share of graphs, ensuring diversity in structural characteristics such as degree distribution, clustering, and connectivity patterns. 

For reproducibility, the following parameter ranges were used during dataset generation. In the ER model, the edge probability \( p \) was sampled from the interval \([\log(n)/n + 0.01, 0.35]\). In the BA model, the number of edges to attach from a new node \( m \) was sampled from \([1, \min(6, n-1)]\). In the WS model, the even neighborhood size \( k \) was drawn from \([2, \min(n-2, 12)]\) and the rewiring probability \(\beta\) from \([0.05, 0.35]\). Each model was sampled uniformly across its respective parameter range, producing connected, non-degenerate graphs suitable for evaluating structural inference capabilities.

Graphs are stored accompanied by metadata that capture key structural properties. Each graph is represented as an edge list, where every line specifies a pair of connected node identifiers. The dataset consists of 1,200 graphs for training and 120 graphs for testing, evenly distributed across the three models. This design provides a well-controlled yet structurally varied benchmark for assessing the generalization and reasoning abilities of lightweight language models in inferring global and local graph parameters.


All generated graphs, along with their metadata and manifest files, constitute the TinyGraphEstimator dataset, one of the key contributions of this work. It is publicly accessible\footnote{https://doi.org/10.5281/zenodo.17387934} to support reproducibility and further research on graph-structured reasoning in language models.

\subsection{Training Setup}

We fine-tune each base model with parameter-efficient supervised instruction tuning using the TRL \texttt{SFTTrainer}~\cite{vonwerra2022trl} and LoRA adapters. Training data consist of paired \emph{prompt}–\emph{completion} examples: the prompt includes the normalized graph (edge list), the fixed target schema, and concise instructions; the completion is the gold JSON with expected fields. To avoid learning spurious prompt tokens and to sharpen output formatting, we use \emph{completion-only loss}. LoRA adapters are applied to attention and MLP projection modules with rank $r{=}32$, $\alpha{=}32$, and dropout $0.05$. We train with sequence length $2048$, batch size $2$ and gradient accumulation $8$, cosine learning-rate schedule with warmup ratio $0.03$, learning rate $2\!\times\!10^{-4}$, and ten epochs over the training set. 

\paragraph{Compute Configuration}
All experiments (training and validation) were executed on a single NVIDIA RTX~3090 GPU (Ampere, 24GB VRAM) in a standard Linux x86\_64 environment.

\subsection{Validation Protocol}
We evaluate model outputs with a schema- and tolerance-aware validation pipeline that mirrors the training prompt format while remaining model-agnostic. For each sample, the script loads the graph data and gold metadata and normalizes it to a canonical textual form. A structured prompt then instructs the model to return \emph{only} a JSON object matching a fixed schema of given fields (density, degree statistics, triangles, clustering, transitivity, path lengths, diameter, chromatic number, global efficiency). Inference is run on a base model or a base model augmented with LoRA adapters (if provided), with JSON-constrained decoding via \texttt{lm-format-enforcer}~\cite{noamgat2023lmformatenforcer} to prevent malformed outputs. The first JSON object in the response is extracted and strictly checked against the schema.

\subsection{Evaluation Metrics}

Performance was evaluated using four standard regression metrics: the normalized root mean squared error based on range (\(\mathrm{NRMSE}_{\text{range}}\)), the symmetric mean absolute percentage error (sMAPE), the \(R^{2}\) Accuracy, and the NRMSE Accuracy. 

\(\mathrm{NRMSE}_{\text{range}}\) represents the root mean squared error normalized by the range of the target variable, providing a scale-independent measure of average deviation between predicted and true values. 
sMAPE quantifies the mean relative difference between predictions and ground truth, computed symmetrically with respect to both magnitudes to balance over- and underestimation. \(R^{2}\) Accuracy expresses the proportion of variance in the target data explained by the model, calculated as \(100 \times \max(0, R^{2})\), while NRMSE Accuracy converts normalized RMSE with respect to the target standard deviation into a percentage scale using \(100 \times (1 - \mathrm{NRMSE}_{\text{std}})\). All accuracy metrics and sMAPE are reported in percentage form for comparability across parameters. Lower \(\mathrm{NRMSE}_{\text{range}}\) and sMAPE values and higher accuracy scores indicate better predictive performance.

\section{\uppercase{Results}}

\begin{table*}[!t]
\centering
\caption{
\textbf{Comparison of \(\mathrm{NRMSE}_{\text{range}}\) and sMAPE (\%) for graph parameter inference.}
Each model is evaluated in plain (zero-shot) and fine-tuned variants. 
Lower values indicate smaller relative errors and therefore better predictions. 
Fine-tuned results are \underline{underlined}, and the best result for each parameter is shown in \textbf{bold}.
}
\label{tab:nrmse_smape_results}
\resizebox{\textwidth}{!}{
\begin{tabular}{lcccccccccccc}
\toprule
\multirow{3}{*}{\textbf{Parameter}} &
\multicolumn{6}{c}{\(\mathbf{NRMSE}_{\text{range}}\)} &
\multicolumn{6}{c}{\textbf{sMAPE (\%)}} \\
\cmidrule(lr){2-7}\cmidrule(lr){8-13}
 & \multicolumn{2}{c}{\textbf{Llama-3.2-3B}} &
   \multicolumn{2}{c}{\textbf{Qwen-2.5-3B}} &
   \multicolumn{2}{c}{\textbf{Phi-4-mini (4B)}} &
   \multicolumn{2}{c}{\textbf{Llama-3.2-3B}} &
   \multicolumn{2}{c}{\textbf{Qwen-2.5-3B}} &
   \multicolumn{2}{c}{\textbf{Phi-4-mini (4B)}} \\
\cmidrule(lr){2-3}\cmidrule(lr){4-5}\cmidrule(lr){6-7}\cmidrule(lr){8-9}\cmidrule(lr){10-11}\cmidrule(lr){12-13}
 & Plain & Fine-tuned & Plain & Fine-tuned & Plain & Fine-tuned &
   Plain & Fine-tuned & Plain & Fine-tuned & Plain & Fine-tuned \\
\midrule
Density                      & 0.902 & \underline{\textbf{0.000}} & 0.317 & \underline{0.005} & 0.998 & \underline{0.014} & 72.764 & \underline{\textbf{0.000}} & 56.069 & \underline{0.201} & 92.402 & \underline{0.654} \\
Degree Min                   & 0.349 & \underline{0.100} & 0.280 & \underline{0.090} & 0.270 & \underline{\textbf{0.088}} & 73.697 & \underline{\textbf{16.717}} & 46.771 & \underline{17.376} & 52.291 & \underline{17.393} \\
Degree Mean                  & 0.397 & \underline{0.001} & 0.311 & \underline{\textbf{0.000}} & 0.160 & \underline{0.004} & 65.803 & \underline{0.011} & 45.800 & \underline{\textbf{0.000}} & 23.579 & \underline{0.217} \\
Degree Max                   & 0.284 & \underline{\textbf{0.033}} & 0.250 & \underline{0.034} & 0.331 & \underline{0.037} & 44.846 & \underline{\textbf{3.335}} & 34.764 & \underline{3.594} & 60.272 & \underline{4.559} \\
Degree Std                   & 0.268 & \underline{0.056} & 0.293 & \underline{0.054} & 0.257 & \underline{\textbf{0.053}} & 48.499 & \underline{\textbf{9.896}} & 44.476 & \underline{9.848} & 45.682 & \underline{10.266} \\
Triangles Total              & 0.314 & \underline{0.030} & 0.198 & \underline{\textbf{0.023}} & 0.291 & \underline{0.037} & 163.371 & \underline{14.027} & 86.373 & \underline{\textbf{9.808}} & 177.232 & \underline{16.806} \\
Average Clustering           & 0.594 & \underline{0.076} & 0.370 & \underline{\textbf{0.056}} & 0.510 & \underline{0.083} & 168.042 & \underline{14.209} & 105.227 & \underline{\textbf{10.034}} & 178.104 & \underline{15.928} \\
Transitivity                 & 0.619 & \underline{0.053} & 0.368 & \underline{\textbf{0.042}} & 0.518 & \underline{0.053} & 168.327 & \underline{10.312} & 102.598 & \underline{\textbf{8.151}} & 178.718 & \underline{11.463} \\
Average Shortest Path Length & 0.237 & \underline{\textbf{0.048}} & 0.222 & \underline{0.051} & 0.345 & \underline{0.058} & 26.490 & \underline{\textbf{2.720}} & 32.143 & \underline{2.794} & 134.941 & \underline{3.286} \\
Diameter                     & 0.215 & \underline{0.068} & 0.236 & \underline{\textbf{0.061}} & 0.218 & \underline{0.074} & 38.764 & \underline{7.399} & 39.536 & \underline{\textbf{5.445}} & 46.950 & \underline{6.565} \\
Chromatic Number             & 0.224 & \underline{0.078} & 0.195 & \underline{0.076} & 0.348 & \underline{\textbf{0.075}} & 26.137 & \underline{5.887} & 21.712 & \underline{\textbf{5.298}} & 41.875 & \underline{5.413} \\
Global Efficiency            & 0.894 & \underline{\textbf{0.017}} & 0.353 & \underline{0.019} & 0.994 & \underline{0.026} & 140.083 & \underline{1.290} & 32.294 & \underline{1.272} & 176.948 & \underline{1.778} \\
\midrule
\textbf{Overall (Mean)}      & 0.441 & \underline{0.047} & 0.283 & \underline{\textbf{0.043}} & 0.437 & \underline{0.050} & 86.402 & \underline{7.150} & 53.980 & \underline{\textbf{6.152}} & 100.750 & \underline{7.861} \\
\bottomrule
\end{tabular}
}
\end{table*}

\begin{table*}[!t]
\centering
\caption{
\textbf{Comparison of models on \(R^{2}\) Accuracy (\%) and NRMSE Accuracy (\%) across graph parameters.}
Higher values indicate better predictive performance and closer agreement between predicted and true graph-structural properties.
Best results for each parameter are shown in \textbf{bold}.
}
\label{tab:param_accuracy_finetuned}
\resizebox{\textwidth}{!}{
\begin{tabular}{lcccccc}
\toprule
\textbf{Parameter} &
\multicolumn{3}{c}{\textbf{R\textsuperscript{2} Accuracy (\%)}} &
\multicolumn{3}{c}{\textbf{NRMSE Accuracy (\%)}} \\
\cmidrule(lr){2-4}\cmidrule(lr){5-7}
 & \textbf{Llama-3.2-3B} & \textbf{Qwen-2.5-3B} & \textbf{Phi-4-mini (4B)} 
 & \textbf{Llama-3.2-3B} & \textbf{Qwen-2.5-3B} & \textbf{Phi-4-mini (4B)} \\
\midrule
Density                      & \textbf{100.000} & 99.957 & 99.621 & \textbf{100.000} & 97.926 & 93.840 \\
Degree Min                   & 85.213 & 87.957 & \textbf{88.415} & 61.546 & 65.297 & \textbf{65.963} \\
Degree Mean                  & 99.999 & \textbf{100.000} & 99.974 & 99.697 & \textbf{100.000} & 98.394 \\
Degree Max                   & \textbf{97.924} & 97.885 & 97.415 & \textbf{85.591} & 85.456 & 83.921 \\
Degree Std                   & 95.286 & 95.599 & \textbf{95.775} & 78.289 & 79.021 & \textbf{79.445} \\
Triangles Total              & 98.284 & \textbf{98.996} & 97.474 & 86.898 & \textbf{89.981} & 84.107 \\
Average Clustering           & 92.539 & \textbf{95.845} & 90.945 & 72.686 & \textbf{79.617} & 69.908 \\
Transitivity                 & 96.465 & \textbf{97.753} & 96.384 & 81.198 & \textbf{85.009} & 80.984 \\
Average Shortest Path Length & \textbf{94.249} & 93.636 & 91.603 & \textbf{76.020} & 74.773 & 71.022 \\
Diameter                     & 88.521 & \textbf{90.875} & 86.226 & 66.120 & \textbf{69.792} & 62.886 \\
Chromatic Number             & 86.986 & 87.487 & \textbf{87.988} & 63.926 & 64.626 & \textbf{65.341} \\
Global Efficiency            & \textbf{99.485} & 99.317 & 98.766 & \textbf{92.822} & 91.736 & 88.893 \\
\midrule
\textbf{Overall (Mean)}      & 94.579 & \textbf{95.442} & 94.216 & 80.339 & \textbf{81.936} & 78.725 \\
\bottomrule
\end{tabular}
}
\end{table*}

\begin{figure*}[!t]
    \centering
    \includegraphics[width=\linewidth]{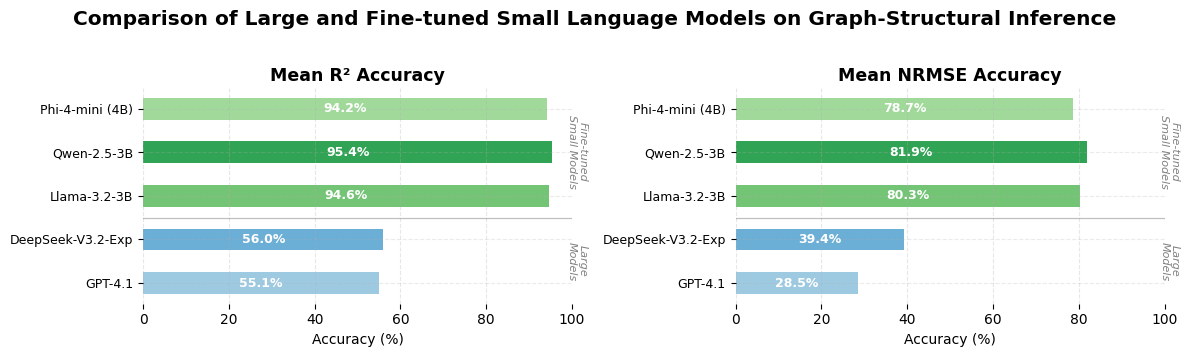}
    \caption{
        \textbf{Comparison of Large and Fine-tuned Small Language Models on Graph-Structural Inference.}
The figure compares average performance across R\textsuperscript{2}~Accuracy and NRMSE~Accuracy.
Compact, fine-tuned small models substantially outperform state-of-the-art large models on graph-structural reasoning tasks, demonstrating the effectiveness of efficient, domain-adapted fine-tuning.
    }
    \label{fig:model_comparison}
\end{figure*}

A comparison of the performance of base and fine-tuned models is presented in Table~\ref{tab:nrmse_smape_results} (accuracy results for untrained models are near zero and therefore omitted). 
The results show a clear and systematic improvement across all evaluated metrics. 
Fine-tuning leads to a substantial reduction in normalized errors, with each model achieving markedly higher precision after adaptation. 
For \(\mathrm{NRMSE}_{\text{range}}\), the mean error decreases from 0.441 to 0.047 for Llama-3.2-3B, from 0.283 to 0.043 for Qwen-2.5-3B, and from 0.437 to 0.050 for Phi-4-mini (4B), corresponding to an average improvement of approximately 0.340. 
A similar trend is observed for sMAPE, where mean values drop from 86.4\% to 7.2\% for Llama, from 54.0\% to 6.2\% for Qwen, and from 100.8\% to 7.9\% for Phi - an overall reduction exceeding 70 percentage points. 
These results demonstrate that LoRA fine-tuning consistently and substantially enhances model precision, reducing both absolute and proportional errors by more than an order of magnitude.

Table~\ref{tab:param_accuracy_finetuned} summarizes the accuracy of the fine-tuned models. 
The results remain consistently very high, with \(R^{2}\) Accuracy exceeding 94\% and NRMSE Accuracy surpassing 78\% across all models and parameters. 
Among them, Qwen-2.5-3B achieved the best overall performance, reaching a mean \(R^{2}\) Accuracy of 95.4\% and an NRMSE Accuracy of 81.9\%. 
Llama-3.2-3B followed closely with 94.6\% and 80.3\%, while Phi-4-mini (4B) achieved 94.2\% and 78.7\%, respectively. 
Although the differences are modest, Qwen shows a consistent advantage on metrics capturing higher-order structural organization, such as transitivity, clustering, and triangle counts - properties that depend on multi-node relationships and global connectivity. 
This suggests that Qwen’s architecture or pretraining corpus may better encode relational and hierarchical dependencies in serialized graph input.

In addition to the fine-tuned small models, we evaluated two state-of-the-art proprietary systems to contextualize the relative performance of our compact architectures. 
As illustrated in Figure~\ref{fig:model_comparison}, the small models substantially outperform these large-scale counterparts across both evaluation metrics. 
While GPT-4.1 and DeepSeek-V3.2-Exp achieve mean~R\textsuperscript{2}~Accuracies of 55.1\% and 56.0\%, and Mean~NRMSE~Accuracies of 28.5\% and 39.4\%, respectively, the fine-tuned small models reach over 94\% and 78\% on average for the same metrics. 
This consistent performance margin underscores the effectiveness of targeted, parameter-efficient adaptation over sheer model scale, demonstrating that compact models can achieve high reasoning precision when optimized for structural inference. 

The fine-tuned small models maintain stable accuracy across diverse graph types - from sparse to dense topologies - demonstrating strong generalization and resilience to variations in degree distribution and connectivity. These outcomes highlight the effectiveness of LoRA-based adaptation in enhancing the structural inference capabilities of compact language models.

\section{\uppercase{Discussion}}


The results highlight how targeted, parameter-efficient adaptation enables small transformer models to acquire reliable structural inference skills from textual graph representations. Consistent improvements across all evaluated graph parameters confirm that small language models are capable of internalizing both local and global structural dependencies without requiring explicit graph neural architectures.

The ability of compact models to infer structural patterns can be attributed to several complementary mechanisms. First, the self-attention mechanism implicitly constructs relational dependencies between tokens, enabling recovery of adjacency-like patterns from serialized edge lists. Second, weight sharing and low-rank adaptation promote compression of relational rules, supporting the discovery of generalizable structural regularities rather than memorization. Third, pretrained linguistic priors - such as token co-occurrence, dependency tracking, and approximate counting - introduce transferable inductive biases that facilitate reasoning over graph-like inputs. Finally, the hierarchical organization of attention layers allows aggregation from local edge information to global structural descriptors, enabling accurate estimation of higher-order metrics such as clustering, transitivity, and global efficiency. Together, these mechanisms explain how small transformers, despite limited capacity and the absence of explicit graph inductive biases, can learn and generalize structural patterns directly from textual graph representations.

These findings have implications for both theory and methodology. They suggest that structural reasoning is not solely an emergent property of model scale but can also be encouraged through targeted adaptation, indicating that compact, efficiency-oriented architectures are capable of meaningful reasoning performance when appropriately fine-tuned. Building on this insight, the TinyGraphEstimator framework provides a controlled and reproducible environment for investigating such behaviors within the domain of graph reasoning, offering a principled way to examine how small language models interpret and generalize structural patterns. Although this study focuses on graph-structured data, the same methodological principles could extend to other structured domains in which relational dependencies are fundamental.

Beyond conceptual insights, these results suggest practical opportunities for deploying small language models in resource-constrained or embedded environments. Fine-tuned compact transformers can serve as efficient reasoning components within larger autonomous or decision-support systems, supporting applications such as knowledge graph analysis, network diagnostics, and symbolic planning. By enabling small models to infer and interpret graph-structural properties effectively, TinyGraphEstimator highlights a pathway toward lightweight reasoning agents that combine interpretability, adaptability, and scalability. Overall, the demonstrated predictive performance and generalization ability of fine-tuned small models help bridge the gap between discrete symbolic reasoning and natural language modeling, establishing compact transformers as viable tools for structured inference across both research and applied contexts.

\section{\uppercase{Limitations}}

While this study demonstrates the potential of small language models for graph-structural reasoning, several limitations should be acknowledged. First, although LoRA fine-tuning consistently improves inference accuracy, the models still exhibit challenges in predicting complex or global parameters such as chromatic number and average shortest path length, which require deeper combinatorial reasoning. Second, the evaluation is based on synthetic graphs generated from a selected set of well-established random graph models, which provide controlled variability but may not reflect the full diversity of real-world network structures. Incorporating additional graph types or domain-specific datasets in future work could further strengthen the assessment of model generalization. Third, the experiments rely on serialized graph formats that, while simple and interpretable, may not capture complex or densely connected structures as effectively as more expressive representations. Future work could explore alternative input forms or hybrid architectures that integrate symbolic and neural reasoning. Despite these limitations, the results highlight promising directions for lightweight, adaptable language models in graph-theoretic inference.

\section{\uppercase{Conclusion}}

This work explored the capability of small language models to infer a wide range of graph-theoretic parameters directly from structured graph representations. Using the proposed TinyGraphEstimator dataset, we systematically evaluated compact transformer-based models on tasks involving both local and global structural reasoning. The results demonstrate that even models with fewer than 4B parameters possess capacity to approximate structural graph measures when adapted through parameter-efficient fine-tuning using the LoRA method. Across all evaluated metrics, fine-tuned models consistently outperformed their zero-shot baselines, confirming that lightweight adaptation can effectively enhance structured reasoning without the computational overhead of large-scale models. 

Overall, this study provides new insight into how small language models can internalize and generalize graph-structural information. By combining efficiency, adaptability, and interpretability, such models offer a promising foundation for scalable graph reasoning and for extending language-based inference to structured, symbolic, and relational domains.

\section*{\uppercase{LLM usage statement}}

This manuscript acknowledges the use of ChatGPT~\cite{chatgpt}, powered by the GPT-5 language model developed by OpenAI, to improve language clarity, refine sentence structure, and enhance overall writing precision.

\bibliographystyle{apalike}
{\small
\bibliography{main}}

\end{document}